\documentclass{article}

\usepackage{microtype}
\usepackage{graphicx}
\usepackage{subfigure}
\usepackage{booktabs} % for professional tables

\usepackage{amsmath}    
\usepackage{amssymb}  
\usepackage{multirow}
\usepackage{placeins}
\usepackage{graphbox}
\usepackage{caption}

\usepackage{hyperref}

\DeclareMathOperator\arctanh{arctanh}
\DeclareMathOperator\softmax{softmax}

\usepackage[accepted]{icml2019}

\icmltitlerunning{On the Vulnerability of Capsule Networks to Adversarial Attacks}

\begin{document}
\twocolumn[
\icmltitle{On the Vulnerability of Capsule Networks to Adversarial Attacks}



\icmlsetsymbol{equal}{*}

\begin{icmlauthorlist}
\icmlauthor{Felix Michels}{equal,hhu}
\icmlauthor{Tobias Uelwer}{equal,hhu}
\icmlauthor{Eric Upschulte}{equal,hhu}
\icmlauthor{Stefan Harmeling}{hhu}
\end{icmlauthorlist}

\icmlaffiliation{hhu}{Department of Computer Science, Heinrich-Heine-Universit\"at D\"usseldorf, Germany}

\icmlcorrespondingauthor{Felix Michels}{felix.michels@hhu.de}
\icmlcorrespondingauthor{Tobias Uelwer}{tobias.uelwer@hhu.de}
\icmlcorrespondingauthor{Eric Upschulte}{eric.upschulte@hhu.de}
\icmlcorrespondingauthor{Stefan Harmeling}{harmeling@hhu.de}

\icmlkeywords{Adversarial Examples, Capsule Networks}

\vskip 0.3in
]



\printAffiliationsAndNotice{\icmlEqualContribution} % otherwise use the standard text.

\begin{abstract}
	This paper extensively evaluates the vulnerability of capsule networks to different adversarial attacks. Recent work suggests that these architectures are more robust towards adversarial attacks than other neural networks. However, our experiments show that capsule networks can be fooled as easily as convolutional neural networks.
\end{abstract}

\section{Introduction}
Adversarial attacks change the input of machine learning models in a
way that the model outputs a wrong result. For neural networks these
attacks were first introduced by Goodfellow et al. \yrcite{fgsm} with
alarming results. Recently capsule networks (CapsNets) \cite{capsules}
have been shown to be a reasonable alternative to convolutional neural
networks (ConvNets). Frosst et al. \yrcite{darccc} state that CapsNets
are more robust against white-box adversarial attacks than other
architectures. Adversarial robustness of CapsNets has been previously
studied by Marchisio et al. \yrcite{marchisio}, but with focus on the
evaluation of their proposed attack. Detecting adversarial examples
using the reconstruction quality of the CapsNets has been investigated
by Frosst et al. \yrcite{darccc}. Also Peer et al. \yrcite{training}
have briefly discussed the application of the fast gradient sign
method (FGSM) \cite{fgsm} on CapsNets. Hinton et al. \yrcite{em}
report results of the FGSM on CapsNets using EM routing.  Another
established approach for CapsNets is the dynamic routing algorithm
\cite{capsules}.

In this paper, we will focus on these variants of CapsNets and
investigate their robustness against common adversarial attacks.  In
particular, we compare the results of four different attacks on
CapsNets trained on different datasets and examine the transferability
of adversarial perturbations. We will show that CapsNets are in
general not more robust to adversarial attacks than ConvNets.

Our paper is structured as followed: in Sec.~\ref{lab:capsules} we
recapitulate the idea of CapsNets and the dynamic routing algorithm. In Sec.~\ref{lab:attacks} we describe the attacks
we apply to CapsNets. We summarize our experiments and results in  Sec.~\ref{lab:experiments}.

\section{Capsule Networks and Dynamic Routing}
\label{lab:capsules}
The concept of vector capsules and the dynamic routing algorithm was proposed by Sabour et al. \yrcite{capsules}. In an essence, neurons are grouped into vectors, so-called capsules. Each capsule vector is dedicated to a distinct abstract entity, i.e. a single object class in a classification setting. The norm of a capsule vector encodes the probability of the represented object being present in the input, while the vector orientation encodes the object's characteristics. Thus, CapsNets aim to develop dedicated representations that are distributed into multiple vectors in contrast to convolutional networks that utilize an entangled representation in a single vector at a given location. This allows the application of linear transformations directly to the representations of respective entities. Spatial relations, which can be implemented as a matrix product, can thus be modeled more efficiently.

CapsNets are organized in layers. Initially, the original CapsNet applies a convolutional layer. The resulting feature maps are then processed by the primary capsule layer. Internally, it applies a series of convolutional layers on its own, each yielding a spatial grid of capsules. Within all capsule layers the \textit{squashing} function serves as a vector-to-vector non-linearity that squashes each capsule vector length between $0$ and $1$ while leaving the orientation unaltered. Subsequently, convolutional or densely connected capsule layers can be applied. While the latter does not utilize weight sharing, convolutional capsule layers share the kernels over the spatial grid, as well as capsules from the previous layer. These layers effectively estimate output capsules based on respective input capsules. The dynamic routing algorithm determines each agreement between estimate and iteratively calculated output capsule. This is done by introducing a scalar factor, the so-called routing coefficient, for each connection between an estimate and respective output. Such an output is defined as the sum over all respective estimates, weighted by their routing coefficients. Theoretically, that means information flows where it is needed, both during forward and backpropagation. This non-parametric procedure supports the goal of capsules with clean dedicated representations. To improve results, an additional capsule may be used within the routing algorithm to serve as a dead end for information that may not be linked to known abstract capsule categories. This is also referred to as the \emph{none-of-the-above} category.

\section{Adversarial Attacks}
\label{lab:attacks}

Adversarial attacks can be performed in different settings: 
\emph{white-box} attacks compute the gradient of the
networks output with respect to the input, whereas in the \emph{black-box}
setting such calculations are not possible. Furthermore, adversarial
attacks can be classified into \emph{targeted} attacks, where the goal of the
attack is that the network assigns a chosen label to the manipulated
image, and \emph{untargeted} attacks, where the attacker's goal is to fool
the network in the sense that it missclassifies a given image.

Throughout this paper we denote the input image as $x\in [0,1]^{n\times n}$, the neural network's output logits as $Z(x)$ and the perturbation as $\delta$. If $F(x)$ is the output of the network interpretable as probability, then
$F(x) = \softmax (Z(x))$ in the case of the ConvNet and $Z(x) = \arctanh(2F(x) - 1)$ in the case of the CapsNet. We refer to the label assigned to $x$ by the networks as $C(x)$  and to the correct label of $x$ by $C^*(x)$. Furthermore, we denote the $i$-th entry of $Z(x)$ as $Z(x)_i$.

\subsection{Carlini-Wagner Attack}

The Carlini-Wagner (CW) attack \yrcite{carlini} is a targeted white-box attack and performed by solving the following constrained optimization problem
\begin{equation}
\begin{aligned}
& \underset{\delta}{\text{minimize}}
& & ||\delta||_2 + c \cdot \max(G(x,\delta,t)-Z(x)_t, -\kappa) \\
& \text{subject to}
& & x+\delta \in [0,1]^{n \times n},
\end{aligned}
\end{equation}

where $G(x,\delta,t) := \max_{i\neq t}(Z(x+\delta)_i)$ and $c>0$. The parameter  $\kappa > 0$ controls the confidence. The optimal value for $c$, i.e. the smallest value, that results in an adversarial example, is found using a binary search. To ensure the box-constraint on $x+\delta$ the authors suggested the following transform of variables 
\begin{equation}
\delta = \frac{1}{2}(\tanh(w)+1)-x,
\end{equation} 
where the $\tanh$-function is applied componentwise. After this
transformation the optimization problem is treated as unconstrained
and can be solved in terms of $w$ using the Adam
optimizer \cite{adam}. Carlini and Wagner \yrcite{carlini} also proposed two
different approaches to handle the box-constraint: projected gradient
descent and clipped gradient descent. For details we refer the reader
to the original work \cite{carlini}.

\subsection{Boundary Attack}

The idea of the boundary attack as introduced by Brendel et
al. \yrcite{boundary} is to sample a perturbation which leads to a
missclassification of the original image $x^{(0)}:=x$. Additionally,
the desired perturbation should have the smallest possible norm. The
initial perturbation $\delta^{(0)}$ is sampled component\-wise from a
uniform distribution $\delta^{(0)}_{ij}\sim \mathcal{U}(0,1)$. Initial
perturbations, which are not missclassified, are rejected. During the
attack adversarial images are constructed iteratively
$x^{(k+1)}:= x^{(k)}+\delta^{(k)}$ by a random walk close to the
decision boundary. During this random walk the following three
conditions are enforced by appropriate scaling and clipping of the
image and the perturbation:
\begin{enumerate}
\item The new image $x^{(k+1)}$ is in the range of a valid image,
  i.e. in $x^{(k+1)}\in [0,1]^{n\times n}$.
\item The proportion of the size of the perturbation $\delta^{(k)}$
  and the distance to the given image is equal to a given parameter
  $\gamma$.
\item The reduction of the distance from the adversarial image to the
  original image $d(x, x^{(k)})-d(x, x^{(k+1)})$ is proportional to
  $d(x, x^{(k)})$ with $\nu>0$.
\end{enumerate}
The parameters $\gamma$ and $\nu$ are adjusted dynamically, similarly to Trust Region methods.

\begin{table*}[h]
	\begin{minipage}{.45\linewidth}
		\caption{Average perturbation norms for each attack and architecture.}
		\vskip 0.15in
		\centering\scalebox{0.85}{
			\begin{tabular}{llcccc}
				\toprule
				Attack & Network       & MNIST & Fashion & SVHN & CIFAR10  \\
				\midrule
				\multirow{2}{*}{CW} & ConvNet & {$1.40$} & $0.51$ & $0.67$ & $0.37$ \\
				& CapsNet            & $1.82$ & {$0.50$} & {$0.60$} & {$0.23$} \\
				\midrule
				\multirow{2}{*}{Boundary} & ConvNet & {$3.07$} & $1.24$ & $2.42$ & $1.38$ \\
				& CapsNet            & $3.26$ & {$0.93$} & {$1.88$} & {$0.72$} \\
				\midrule
				\multirow{2}{*}{DeepFool} & ConvNet & {$1.07$} & {$0.31$} & {$0.41$} & $0.23$ \\
				& CapsNet           & $2.02$ & $0.55$ & $0.80$ & {$0.16$} \\
				\midrule
				\multirow{2}{*}{Universal} & ConvNet & {$6.71$} & {$2.61$} & {$2.46$} & {$2.45$} \\
				& CapsNet           & $11.45$ & $5.31$ & $8.59$ & $2.70$ \\
				\bottomrule\\
		\end{tabular}}
		\label{tab:norms}
	\end{minipage}\hspace{0.8cm}
	\begin{minipage}{.45\linewidth}
		\caption{Fooling rates of adversarial examples calculated for a CapsNet and evaluated on a ConvNet and vice versa. For the universal attack we report the accuracy on the whole test set.}
		\vskip 0.15in
		\centering\scalebox{0.85}{
			\begin{tabular}{llcccc}
				\toprule
				Attack & Network       & MNIST & Fashion & SVHN & CIFAR10  \\
				\midrule
				\multirow{2}{*}{CW} & ConvNet & $0.8\%$ & $1.2\%$ & $2.8\%$ & $2.4\%$ \\
				& CapsNet            & $2.0\%$ & $2.0\%$ & $3.8\%$ & $2.0\%$ \\
				\midrule
				\multirow{2}{*}{Boundary} & ConvNet & $8.8\%$ & $9.5\%$ & $10.5\%$ & $13.4\%$ \\
				& CapsNet            & $14.2\%$ & $14.6\%$ & $12.9\%$ & $26.1\%$ \\
				\midrule
				\multirow{2}{*}{DeepFool} & ConvNet & $4.3\%$ & $8.5\%$ & $13.5\%$ & $11.8\%$ \\
				& CapsNet           & $0.9\%$ & $10.9\%$ & $10.8\%$ & $14.1\%$ \\
				\midrule
				\multirow{2}{*}{Universal} & ConvNet & $4.9\%$ & $20.4\%$ & $35.0\%$ & $25.9\%$ \\
				& CapsNet           & $38.2\%$ & $25.7\%$ & $53.4\%$ & $47.2\%$ \\
				\bottomrule\\
		\end{tabular}}
		\label{tab:attacks}\end{minipage}
\end{table*}

\subsection{DeepFool Attack}
DeepFool is an untargeted white-box attack developed by
Moosavi-Dezfooli et al. \yrcite{deepfool}.  The authors found that
minimal adversarial perturbations for affine multiclass classifiers can
be computed exactly and quickly, by calculating the distance to the
(linear) decision boundaries and making an orthogonal projection to
the nearest boundary.  DeepFool initializes $\delta^{(0)} \gets 0$ and then
iteratively approximates $F$ with its first degree Taylor polynomial
at $x + \delta^{(k)}$, computes a perturbation $\Delta \delta^{(k)}$
for this approximation as described above and updates
$\delta^{(k+1)} \gets \delta^{(k)} + \Delta \delta^{(k)} $.  For
better results, we restrict the norm of $\Delta \delta^{(k)} $ each
step $k$.

\subsection{Universal Adversarial Perturbations}
\label{sec:universal}
A universal perturbation is a single vector $\delta \in \mathbb{R}^{n\times n}$, such that $C(x + \delta) \neq C^*(x)$ for multiple $x$ sampled from the input image distribution. This concept was proposed by Moosavi-Dezfooli et al. \yrcite{universal} and we use a variation of their algorithm, which we briefly describe in the following. As long as the accuracy on the test set is above a previously chosen threshold, repeat these steps:
\begin{enumerate}
	\item Initialize $\delta^{(0)} \gets 0$.
	\item Sample a batch $X^{(k)} = \{x_1^{(k)}, ..., x_N^{(k)}\}$ of images with $\forall x \in\ X^{(k)}:  C(x + \delta^{(k)}) = C^*(x)$.
	\item For each $x_i^{(k)}$ compute a perturbation $\delta_i^{(k+1)}$ using FGSM \cite{fgsm}.
	\item Update the perturbation: $$\delta^{(k+1)} \gets \delta^{(k)} + \frac{1}{N} \sum\limits_{i=0}^N \delta_i^{(k+1)}$$
	%\item Stop, once a sufficiently low accuracy on the test set is achieved.
\end{enumerate}
Since this method depends on the FGSM it is a white-box attack.

\section{Experiments}
\label{lab:experiments}

\subsection{Datasets and Network Architectures}

We train models on each of the following benchmark datasets: MNIST
\cite{mnist}, Fashion-MNIST \cite{fashion}, SVHN \cite{svhn} and
CIFAR-10 \cite{cifar}. Each dataset consists of ten different
classes. As a baseline architecture we use a ConvNet which we trained
on each of the datasets using batch-normalization
\cite{batchnorm} and dropout \cite{dropout}. Since training CapsNets
can be rather difficult in practice, we had to carefully select
appropriate architectures:

Like Sabour et al. \yrcite{capsules} we use a three layer CapsNet for
the MNIST dataset, where we only used $64$ convolutional kernels in
the first layer. For the Fashion-MNIST and the SVHN dataset we use two
convolutional layers at the beginning ($32$ and $32$ channels with
$3\times3$ filter for Fashion-MNIST, $64$ and $256$ channels with
$5\times5$ filter for SVHN), followed by a convolutional capsule layer
with $8$D capsules, $32$ filter with size $9\times9$ and a stride of
$2$, and finally a capsule layer with one 16D capsule per class. Since
CapsNets have problems with more complex data like CIFAR10
\cite{complex}, we use a modified DCNet \cite{denseanddiverse} with
three convolutional capsule layers and so-called none-of-the-above category for
the dynamic routing for this dataset. We train all CapsNet
architectures using the margin loss and the reconstruction loss for
regularization \cite{capsules}.

For each dataset we calculate $1000$ adversarial examples on images
randomly chosen from the test set using the DeepFool attack and the
boundary attack.  For the Carlini-Wagner attack we calculate $500$
adversarial examples again on random samples from the test set (with
hyperparameter $\kappa = 1$). The target labels too are chosen at
random, but different from the true labels. To evaluate the
performance of universal perturbation we split the test set in ten
parts and compute ten adversarial perturbations according to the
procedure described in Sec.~\ref{sec:universal} on each part.

None of the attacks restrict the norm of the perturbation. This means, the Carlini-Wagner, the boundary and the DeepFool attack generate only valid adversarial examples. In the case of the universal perturbation, we stop once accuracy falls below $50\%$.

\subsection{Results}

We are aware of the fact that the test accuracies shown in Tab.~\ref{tab:accuracies} of our models are not state-of-the-art. However, we found our models to be suitable for the given task, since the similar performances of ConvNets and CapsNets ensure comparability. 
\begin{table}[h]
	\caption{Test accuracies achieved by our networks.}
	\vskip 0.15in
	\centering\scalebox{0.85}{
		\begin{tabular}{lcccc}
			\toprule
			Network       & MNIST & Fashion-MNIST & SVHN & CIFAR10  \\
			\midrule
			ConvNet           & $99.39\%$ & $92.90\%$ & $92.57\%$ & $88.22\%$ \\
			CapsNet           & $99.40\%$ & $92.65\%$ & $92.35\%$ & $88.21\%$ \\
			\bottomrule\\
	\end{tabular}}
	\label{tab:accuracies}
\end{table}

We also compare the average Euclidean norm of the perturbation for each attack, dataset and network. The results are displayed in Tab.~\ref{tab:norms}. Our main result is that applying the Carlini-Wagner attack on the CapsNets yields smaller adversarial perturbations than on the ConvNet. Nevertheless, for most of the dataset we found that the DeepFool attack performs worse on the CapsNets.

To compare the transferability of adversarial examples we calculate
perturbations on the ConvNet and apply those to the CapsNet and vice
versa (see Tab.~\ref{tab:attacks}). In case of the (targeted)
Carlini-Wagner (CW) attack we define a network \emph{fooled} if the
perturbed image is classified with the target label. For the
Carlini-Wagner attack, the boundary attack and the DeepFool attack our
results fit to those displayed in Tab.~\ref{tab:norms}. Especially the
perturbations calculated using the universal attack seem to generalize
well on the other architecture. For this attack we also found out that
the smaller perturbations calculated on the ConvNet can be
successfully transferred to CapsNets, while the other way around this
approach was less effective, although the norms of the perturbations
for the CapsNets are very large.

\begin{figure}[h]
	\centering
	\begin{tabular}{rlll} 
		CW & \includegraphics[height=1.5cm, align=c]{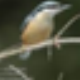} & \includegraphics[height=1.5cm, align=c]{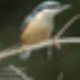} & \includegraphics[height=1.5cm, align=c]{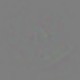}\\
		\\
		Boundary & \includegraphics[height=1.5cm, align=c]{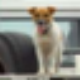} & \includegraphics[height=1.5cm, align=c]{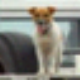} & \includegraphics[height=1.5cm, align=c]{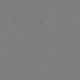}\\
		\\
		DeepFool & \includegraphics[height=1.5cm, align=c]{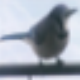} & \includegraphics[height=1.5cm, align=c]{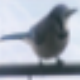} & \includegraphics[height=1.5cm, align=c]{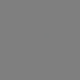}\\
		\\
		Universal & \includegraphics[height=1.5cm, align=c]{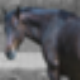} & \includegraphics[height=1.5cm, align=c]{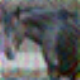} & \includegraphics[height=1.5cm, align=c]{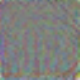}\\
		\\
		\vspace{0.1cm}\\
	\end{tabular}
	\caption{Original images from the CIFAR10 dataset (left), adversarial images (middle) and the corresponding perturbation (right) calculated for a CapsNet.\label{tab:images}}
\end{figure}

The adversarial examples for the CapsNets calculated with the
Carlini-Wagner, the boundary and the DeepFool attack are not
visible for the human eye. Only the universal perturbations are
observable (see Fig.~\ref{tab:images}).

\subsection{Visualizing Universal Perturbations}

\begin{figure}[h]
	\centering
	\includegraphics[height=5.5cm]{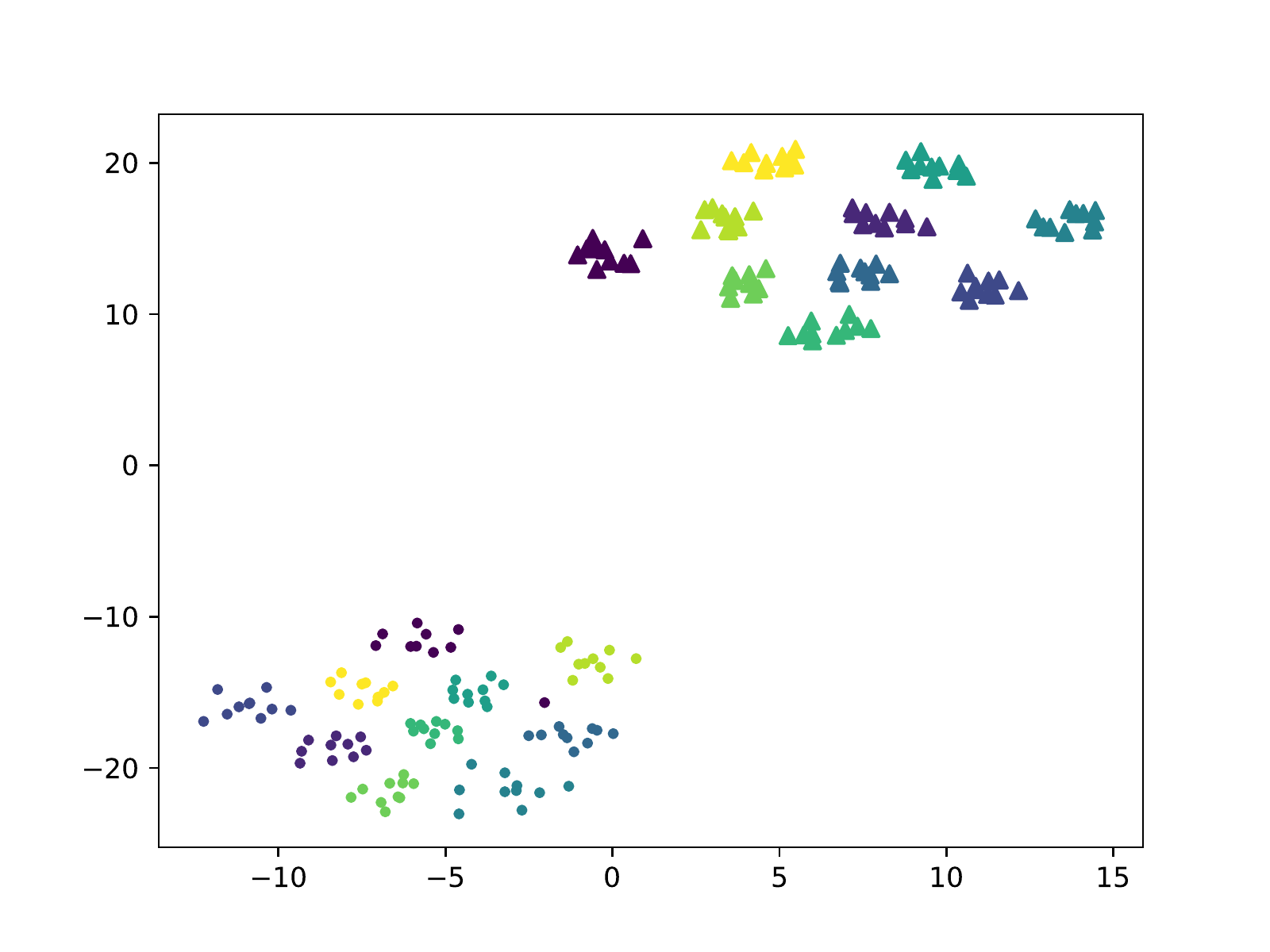}
	\caption{Two dimensional embedding of the universal perturbations calculated using t-SNE \cite{tsne}. The upper right cluster represents perturbations calculated on a ConvNet, whereas the lower left cluster represents those calculated on a CapsNet. Perturbations with the same color were created using the same subset of test data.}
	\label{fig:tsne}
\end{figure}

We also visualized the universal perturbations calculated for the
CapsNet and for the ConvNet using t-SNE \cite{tsne} and we observe
that the perturbations for the CapsNet seem to be inherently different
than the perturbations for the ConvNets (see Fig.~\ref{fig:tsne}).

\section{Conclusion}
Our experiments show that CapsNets are not in general more robust to
white-box attacks. With sufficiently sophisticated attacks CapsNets
can be fooled as easily as ConvNets.  Our experiments also show that
the vulnerability of CapsNets and ConvNets is similar and it is hard
to decide which model is more prone to adversarial attacks than the
other. Moreover, we showed that adversarial examples can be transferred
between the two architectures.

To fully understand the possibly distinguishable roles of the convolutional and capsule layers with respect to adversarial attacks, we are currently examining the effects of attacks on the activation level of single neurons.  However, this analysis is not finished yet and beyond the scope of this paper.

\bibliography{icml}
\bibliographystyle{icml2019}

\end{document}